\documentclass[sigconf,10pt]{acmart}
\usepackage{microtype}
\usepackage{graphicx}
\usepackage{subfigure}
\usepackage{booktabs} 
\usepackage{algorithm}
\usepackage{amsmath}
\usepackage{algpseudocode}
\usepackage{algorithm}
\usepackage{algorithmicx}
\usepackage{algpseudocode}
\usepackage{verbatim}
\usepackage{epstopdf}
\usepackage{multirow}
\usepackage{amsfonts}
\usepackage{soul}
\usepackage{natbib}
\setcitestyle{sort, numbers}
\AtBeginDocument{%
  }

\begin{document}
\settopmatter{printacmref=false} 
\renewcommand\footnotetextcopyrightpermission[1]{} 
\pagestyle{plain} 


\title{MobiLLM: Enabling LLM Fine-Tuning on the Mobile Device via Server Assisted Side Tuning}


\author{Liang Li}
\authornote{Both authors contributed equally to this research.}
\email{lil03@pcl.ac.cn}
\affiliation{%
  \institution{Pengcheng Laboratory}
  \city{Shenzhen}
  \country{China}}

\author{Xingke Yang}
\authornotemark[1]
\email{xyang56@cougarnet.uh.edu}
\affiliation{%
  \institution{University of Houston}
  \city{Houston}
  \country{USA}}

\author{Wen Wu}
\email{wuw02@pcl.ac.cn}
\affiliation{%
  \institution{Pengcheng Laboratory}
  \city{Shenzhen}
  \country{China}}

\author{Hao Wang}
\email{hwang9@stevens.edu}
\affiliation{%
  \institution{Stevens Institue of Technology}
  \city{Hoboken}
  \country{USA}}

\author{Tomoaki Ohtsuki}
\email{ohtsuki@keio.jp}
\affiliation{%
  \institution{Keio University}
  \city{Tokyo}
  \country{Japan}}

\author{Xin Fu}
\email{xfu8@central.uh.edu}
\affiliation{
  \institution{University of Houston}
  \city{Houston}
  \country{USA}}

\author{Miao Pan}
\email{mpan2@uh.edu}
\affiliation{
  \institution{University of Houston}
  \city{Houston}
  \country{USA}}

\author{Xuemin Shen}
\email{sshen@uwaterloo.ca}
\affiliation{
  \institution{University of Waterloo}
  \city{Waterloo}
  \country{Canada}}


\begin{abstract}

Large Language Model (LLM) at mobile devices and its potential applications never fail to fascinate. However, on-device LLM fine-tuning poses great challenges due to extremely high memory requirements and slow training speeds. Even with parameter-efficient fine-tuning (PEFT) methods that update only a small subset of parameters, resource-constrained mobile devices cannot afford them. 
In this paper, we propose MobiLLM to enable memory-efficient transformer LLM fine-tuning on a mobile device via server-assisted side-tuning. Particularly, MobiLLM allows the resource-constrained mobile device to retain merely a frozen backbone model, while offloading the memory and computation-intensive backpropagation of a trainable side-network to a high-performance server. Unlike existing fine-tuning methods that keep trainable parameters inside the frozen backbone, MobiLLM separates a set of parallel adapters from the backbone to create a backpropagation bypass, involving only one-way activation transfers from the mobile device to the server with low-width quantization during forward propagation. In this way, the data never leaves the mobile device while the device can remove backpropagation through the local backbone model and its forward propagation can be paralyzed with the server-side execution. Thus, MobiLLM preserves data privacy while significantly reducing the memory and computational burdens for LLM fine-tuning. Through extensive experiments, we demonstrate that MobiLLM can enable a resource-constrained mobile device, even a CPU-only one, to fine-tune LLMs and significantly reduce convergence time and memory usage.

\end{abstract}

\ccsdesc[500]{Computing methodologies~Distributed artificial intelligence}

\keywords{Large language model, Transformer, Fine-tuning, Memory, Mobile device.}

\maketitle

\section{Introduction}\label{sec:introduction}

Breakthroughs in transformer-based large language models (LLMs), like BERT~\cite{devlin2018bert}, LLaMa ~\cite{touvron2023llama}, and GPT~\cite{brown2020language}, have brought about significant progress in artificial intelligence (AI). Their ability to understand context and nuances allows them to effectively handle various tasks in natural language processing (NLP), computer vision (CV), etc. While LLMs are usually trained on large datasets to offer emergent generalization ability, the continuous generations from private and inaccessible personal data on mobile devices, such as user reviews and messages, often diverge from publicly pre-trained LLM distributions. That necessitates on-device post-deployment fine-tuning to tailor personalized models while ensuring that data remains on the mobile device. 

Advances in hardware are also accelerating the feasibility of this vision. Modern mobile devices, such as the NVIDIA AGX, iPhone 16, and MacBook Pro, have steadily gained powerful on-device computing capabilities optimized for AI workloads~\cite{Li2021infocom,rui2023EEFL}. Recently, there have been great hardware efforts specifically targeting LLM applications on edging mobile or IoT devices. For example, Qualcomm's Snapdragon 8 Elite can execute LLMs at remarkable speeds of 70 tokens per second~\cite{Snapdragon8}, highlighting the potential for real-time on-device LLM tasks. Google is also revolutionizing its smart home ecosystem by integrating Gemini AI---a multi-modal LLM developed by DeepMind \cite{gemini}---into Google Assistant, Nest devices, and the Google Home APP. Those innovations empower users with promising tools for creativity, information summarization, image generation, and smarter interactions through text, voice, and photos~\cite{geminihelp}. Overall, the rapid evolution of LLMs and mobile hardware is driving the development of more intelligent and personalized mobile applications, paving the way for continuous LLM fine-tuning at mobile devices.

Despite those attractive promising applications, the ever-growing contemporary LLMs are actually difficult to fine-tune. The huge gap between the tight resources of mobile devices and the extensive resource demands of on-device LLM fine-tuning makes it even more challenging. Recent research on parameter-efficient fine-tuning (PEFT) aims to reduce trainable parameters by freezing the pre-trained backbone and inserting small trainable modules at various points within it, while PEFT is still too much to afford for a mobile device. Besides, some pioneering approaches explore multi-device cooperative training, which distributes the resource-intensive LLMs fine-tuning tasks across connected mobile devices to alleviate the computational and memory burdens on the individual device. For instance, the pipeline training paradigm partitions an LLM into smaller sub-models, which are concatenated and deployed on cooperative mobile devices~\cite{borzunov2024distributed, lin2023pushing, wang2024federated}. Forward and backward propagations are then performed in a multi-hop relay manner across participating mobile devices. To accelerate such pipeline training, input data are partitioned into micro-batches, which can concurrently be processed across mobile devices~\cite{narayanan2019pipedream, chen2023ftpipehd, ye2024asteroid}.

However, existing approaches cannot support LLM fine-tuning on one ``single" mobile device and may be difficult to deploy in practice. We identify three key obstacles as follows: 1) Existing collaborative LLM fine-tuning schemes typically rely on layer-wise model partitioning and cross-device pipeline training, which requires all devices to be within a local area network with stable peer-to-peer links. It is, however, often impractical in edge networks with limited capable mobile devices and vulnerable to single-device failures, not to mention low delivery ratios and high error rates of multi-hop transmissions.
2) Training LLMs demands enormous memory usage, which often exceeds the capacity of mobile devices (e.g., >70 GB for OPT-6.7B model vs. typical 4GB–12GB DRAM of mobile devices). Although PEFT methods help to reduce memory footprint up to 30\%~\cite{liao2024make}, they cannot fundamentally resolve memory issues since gradients for backpropagation still require passing through the entire pre-trained model. That makes PEFTs inadequate for enabling LLM fine-tuning on the memory constrained mobile device.
3) Fine-tuning process typically follows an alternating forward and backward propagation policy, where backward propagation is more computation-intensive. On-device accelerators like Google Edge TPU or Qualcomm Hexagon are typically optimized for inference (i.e., forward propagation only) and lack acceleration support for backpropagation-specific operations. That may significantly prolong training time or even disable fine-tuning on resource-constrained mobile devices.

In this paper, we introduce MobiLLM, a novel system framework and corresponding implementations, to enable LLM fine-tuning on one ``single" mobile device via server assisted side-tuning. Inspired by side-tuning~\cite{sung2022lst}, the basic idea of MobiLLM is to (i) decouple LLM fine-tuning into the frozen pre-trained modules and a side network consisting of trainable modules, (ii) split the LLM fine-tuning workloads between the mobile device and the server, i.e., the frozen pre-trained backbone model to the mobile device and the trainable side network to the server, and (iii) let inter-layer activations exchange between the mobile device and the server. In this way, MobiLLM keeps memory- and computation-efficient forward pass as well as the functional LLM backbone on the resource-constrained mobile device, and offloads computation-expensive and memory-hungry backpropagation to the high performance computing server, while ensuring the data never leaves the local mobile device.

To the best of our knowledge, MobiLLM is the first work to enable LLM fine-tuning on a mobile device, even a CPU-only one, with the help of the server, which leverages the existing mobile edge network architecture to execute the forward and backward propagation decoupling, and breaks memory and computation barriers of mobile devices. Our contributions are summarized as follows:

\begin{itemize}
	\item We propose MobiLLM, a novel on-device LLM fine-tuning framework. MobiLLM offloads dominant computational and memory burdens to the high-performance server, and allows the resource-constrained mobile device to handle only a frozen backbone model and mobile-friendly operations that fit their hardware capacity. That enables the mobile device to fine-tune LLMs while keeping their data locally on the device. 

    \item We develop a quantized adapter side-tuning method tailored for MobiLLM fine-tuning systems. By stacking a set of parallel adapter modules, a trainable side-network is decoupled from the frozen pre-trained LLM backbone, creating a backpropagation bypass. It requires only one-way activation transfer during the forward pass, utilizing low-precision quantization, which allows the side-network to be trained by leveraging the intermediate representations of the pre-trained model.

    	\item We deploy MobiLLM on several popular mobile devices or development kits, including NVIDIA Xavier and CPU-only laptops, and evaluate its performance using both sub-billion-sized LLM (OPT-350M) and billion-sized LLM (OPT-1.3B), across various tasks and system configurations. 
 The results demonstrate that MobiLLM enables billion-sized LLM fine-tuning on memory constrained mobile devices like Xavier (with 4.6 GB available GPU RAM only). Compared with SOTA methods, MobiLLM remarkably reduces its memory usage up to $4\times$ and the convergence time up to $2.3\times$ for on-device LLM fine-tuning at limited communication costs.

\end{itemize}

The remainder of the paper is organized as follows. Section II analyzes the limitations of existing work and introduces our motivation. Section III elaborates on our MobiLLM design. Section IV introduces the MobiLLM implementation. Section V presents the performance evaluation, and Section VI finally concludes the paper.

\section{Background and Motivation}\label{sec:motivation}

\subsection{Parameter-efficient Fine-Tuning}

PEFT is a feasible direction that adapts a pre-trained model to a downstream task by only training a few trainable parameters, instead of updating all parameters. A popular line of work on PEFT is to add a few trainable parameters and only tune them. For example, Adapters~\cite{houlsby2019adapter} are small bottleneck modules that are inserted into transformer layers, and experiments have shown that training adapters with layer normalization layers are sufficient to achieve full fine-tuning performances. In a similar trend, LoRA~\cite{hulora} injects trainable rank decomposition matrices into a frozen pre-trained model. Instead of inserting new parameters into pre-trained models, prompt-based methods~\cite{zhao2023fedprompt} add trainable parameters to the input and keep the entire pre-trained model unchanged during training. The inserted prompts learn to make use of the knowledge of the pre-trained model to solve new tasks. In addition to approaches that introduce new parameters, there are also various methods that select a sparse subset of parameters from the pre-trained model to update, without adding any new parameters. One such representative method is BitFit~\cite{zaken2021bitfit}, which updates every bias term in the model. While PEFT methods aim to achieve competitive results with minimal parameter updates, they still require significant GPU memory and can be slow during the fine-tuning process. This is because the updated parameters are inside the pre-trained large model, necessitating a full backward pass through the model to compute gradients for backpropagation. This prevents PEFT methods from being directly applied to many real-world applications with limited computational resources, and thus does not help to fit for fine-tuning a large language model on a GPU of a resource-constrained mobile device.

\subsection{LLM Fine-tuning on Mobile Devices}

On-device fine-tuning of personal data is a powerful solution for adapting LLMs to user-specific tasks while maintaining privacy, as all data storage and computation occur locally without any data leaving the device. To fit an LLM on a mobile device, there are two major techniques: \textit{model compression} and \textit{collaborative fine-tuning}. 
Model compression compresses a backbone network to a smaller one, making it more manageable for training. Techniques like pruning and quantization reduce the number of weights and the bit-width needed to represent these weights, respectively~\cite{han2015deep}. 
However, excessive compression can lead to significant drops in performance, necessitating a careful balance between resource savings and model accuracy. 
Collaborative fine-tuning employs devices with fine-tuning tasks to discover nearby nodes with devices handling fine-tuning tasks discover nearby nodes capable of establishing direct communication links and share the computational and memory load of the fine-tuning process. 
This is often based on exchanging necessary intermediate parameters. For example, multi-device pipeline parallelism splits the LLM by layers into smaller sub-models that are deployed across different devices, where forward and backward propagation computations are performed in a relay fashion across these devices~\cite{borzunov2024distributed,wang2024federated}. This pipeline can be further accelerated by partitioning input data into micro-batches and feeding them continuously across devices to enhance parallelism~\cite{narayanan2019pipedream,ye2024asteroid}. This collaborative fine-tuning paradigm effectively leverages the computational and transmission capabilities of neighboring nodes within a network, allowing flexible task deployment and execution under individual computational and memory constraints.

\subsection{Motivation}


In this subsection, we reveal three key issues faced by practical on-device LLM fine-tuning:

\textbf{Observation-1: Limitations of cross-device pipeline fine-tuning.} Most of SOTA cooperative LLM fine-tuning schemes rely on layer-wise model partitioning and cross-device pipeline training, which typically require stable end-to-end multi-hop transmission links among participating mobile devices within a local area network. However, in real-world networks, there may not be enough capable mobile devices to form a stable collaboration loop, and single-device failures in the chain can disrupt the system or even ruin the overall LLM fine-tuning. Besides, multi-hop transmissions are notorious for the low delivery ratio and high error rates. 
Even assuming that the pipeline fine-tuning works well, its ``relay-style'' training process inevitably introduces waiting time at each participating mobile device (awaiting intermediate results from the preceding device in the pipeline). While some parallel training methods may mitigate devices' idle time by using stale model parameters for concurrent computation, this can impair model convergence and incur significant coordination overhead. Furthermore, since each device in the pipeline maintains only a local portion of the LLM, individual mobile device cannot employ the entire fine-tuned model to perform inference independently, which hinders local on-device LLM service provisioning. 

\begin{figure}
    \centering
    \includegraphics[width=0.85\linewidth]{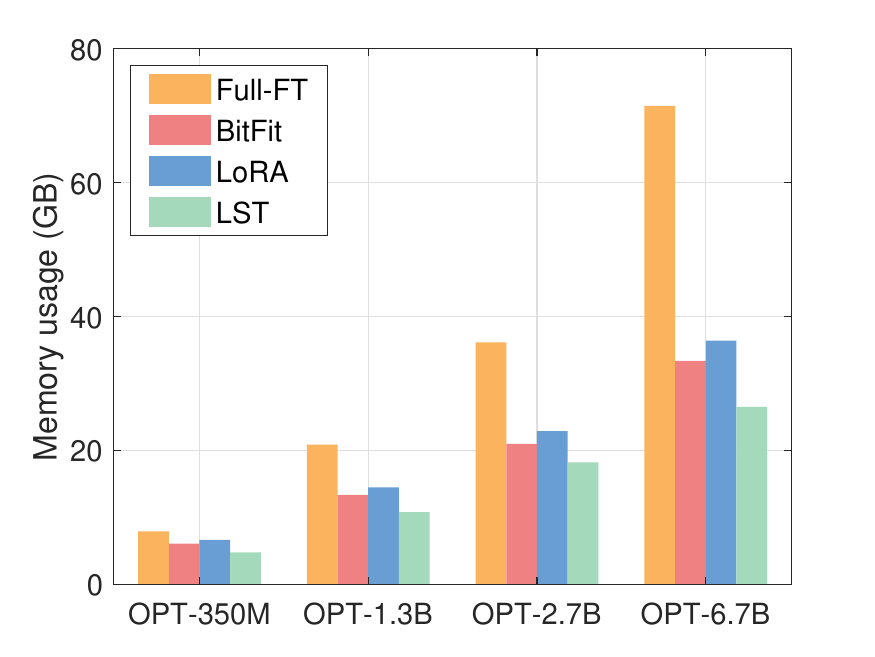}
    \caption{Peak memory footprint of different methods.}
    \label{fig:memoryfootprint}
\end{figure}

\begin{table}[]
\scalebox{1}{
    \begin{tabular}{cccccc}
\hline
\multirow{2}{*}{\textbf{Method}} &
  \multicolumn{1}{c|}{\multirow{2}{*}{\textbf{\begin{tabular}[c]{@{}c@{}}Trainable \\ Para.\end{tabular}}}} &
  \multicolumn{4}{c}{\textbf{Memory Footprint (GB)}} \\ \cline{3-6} 
 &
  \multicolumn{1}{c|}{} &
  \multicolumn{1}{c|}{\textbf{Model}} &
  \multicolumn{1}{l|}{\textbf{Act.}} &
  \multicolumn{1}{l|}{\textbf{Opt.}} &
  \textbf{Total} \\ \hline
Full-FT   & 1.3B   & 2.509 & 10.859 & 7.527 & 20.895 \\
BitFit    & 0.52M  & 2.509 & 10.859 & 0.003 & 13.371 \\
LoRA      & 12.35M & 2.533 & 11.964 & 0.072 & 14.569 \\ \hline
Inference & ---       & 2.509 & 1.894  & ---      & 4.403  \\ \hline
\end{tabular}
}
\caption{The breakdown of memory footprint. (Model: OPT-1.3B; batch size: 16; sequence length: 256; Full-FT: full fine-tuning.)} \label{mot-membreak}
\end{table}

\textbf{Observation-2: Challenges imposed by memory bottlenecks.} Figure~\ref{fig:memoryfootprint} shows the peak memory usage in training various LLMs with a batch size of 16. Table~\ref{mot-membreak} further provides a detailed breakdown of the memory footprint for fine-tuning an OPT-1.3B model. The observed memory requirements are often unaffordable for mobile devices (e.g., more than 70 GB for OPT-6.7B model vs. typical 4GB–12GB DRAM of mobile devices). For smartphones, this situation is even worse, since only a fraction of the memory can be allocated to training tasks without affecting the user's experiences. In contrast, inference (forward propagation only) consumes much less memory (e.g., 4.403GB) because it does not need to store intermediate activations for all layers, which are necessary for backpropagation during training. In fine-tuning, those activations take up most of the memory, and this memory usage quickly scales up with batch size, sequence length, hidden layer dimensions, and the number of layers. Consequently, given the fact that LLMs typically have large dimensions and numerous layers, a slight increase in sequence length may lead to memory overflow for mobile devices, which can support LLM fine-tuning with smaller sequence-length data samples. Those harsh memory challenges result in unreliable on-device LLM fine-tuning, where tasks may occasionally succeed or fail. It is also difficult to assess whether a particular mobile device can support local LLM fine-tuning.

\textbf{Observation-3: Inefficiency of on-device backpropagation computing.} Fine-tuning process typically follows an alternating forward and backward propagation policy, where backward propagation is more computation-intensive --- roughly twice the cost of forward propagation. While pipeline parallel training \cite{narayanan2019pipedream} enables multi-batch parallel processing, concurrent forward and backward computation on each device can lead to competition for computational resources. These factors can severely prolong processing time or even cause task failure on resource constrained mobile devices. Advanced mobile devices may be equipped with powerful and fast-evolving DNN accelerators (NPUs), e.g., Google Edge TPU and Qualcomm Hexagon, but these accelerators are tailored for inference rather than training. They lack good support for backpropagation-specific operations like dynamic gradient updating~\cite{xu2023FwdLLM}, which limits their accelerating performance for LLM fine-tuning tasks.

In summary, current mobile devices are incapable of handling LLM fine-tuning tasks due to the considerably high demands on hardware resources. Simply freezing pre-trained parameters or decentralizing tuning tasks offers limited improvement. To achieve practical and resource-efficient LLM fine-tuning on the mobile device, it requires a holistic solution jointly considering model architecture, training paradigms, and network support.

\section{MobiLLM Design}
\subsection{MobiLLM Overview}
MobiLLM is a mobile device-server framework designed to enable memory-efficient LLM fine-tuning on the mobile device. The key idea is to create a backpropagation bypass parallel to the frozen backbone LLM, and split the LLM fine-tuning between the mobile device (the fixed forwardpass) and the server (the memory and computation intensive backpropagation), while keeping local data on the device.

\begin{figure*}
    \centering
    \includegraphics[width=0.97\textwidth]{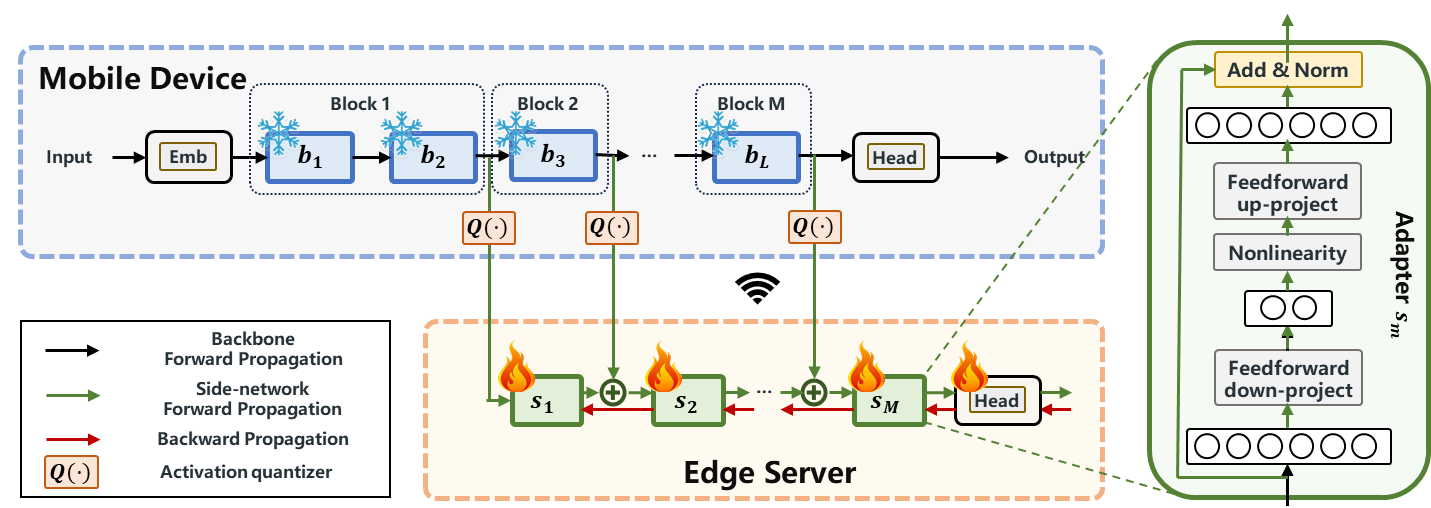}
    \caption{An overview of the MobiLLM system. }
    \label{fig:overview}
\end{figure*}

Figure~\ref{fig:overview} presents an overview of the MobiLLM system. MobiLLM allows the mobile device to retain the local training dataset and pre-trained LLM backbone. Meanwhile, a side-network is deployed and trained on the server connected to the mobile device via stable wireless transmission links. During the fine-tuning, the mobile device performs only forward propagation on the backbone model, where intermediate activations are extracted and served as the inputs for the side-network via shortcut connections. In this way, MobiLLM completely separates the mobile device from those expensive backpropagation computations and saves substantial on-device memory (intermediate activations and optimizer states) during LLM fine-tuning. As a potential cost, the MobiLLM framework introduces some communication overheads and transmission latencies due to the transfer of activation values from the mobile device to the server. To tackle those issues, we provide a communication-efficient side-tuning method using activation quantization (Sec.~\ref{subsec:Quantized Adapter Side-tuning}) and optimize the device-server training procedures (Sec.~\ref{sec:procedure}) to effectively reduce transmission latencies in the proposed MobiLLM system, ensuring memory-friendly, computation-light and time-efficient LLM fine-tuning on the mobile device. 

\subsection{Quantized Adapter Side-tuning}
\label{subsec:Quantized Adapter Side-tuning}	
In Fig.~~\ref{fig:overview}, the blue boxes are the original transformer-based LLM backbone, and the green boxes outline our side-network that is composed of a set of trainable modules. Specifically, we separate the trainable modules from the backbone network rather than plug them in, thus providing a dedicated ``highway’’ where all trainable parameters are on this highway. The remainder of this subsection elaborates on the design of the side-network and its connection with the backbone.

	\textbf{Parallel adapter based side-network.} MobiLLM establishes a side-network parallel to the main backbone, which refines the backbone output to produce more accurate representations. In particular, we design our side-network by using stacked parallel adapter modules. The input to each adapter combines the intermediate activation output of the transformer layer in the backbone and the activation of the previous adapter. This ensures that our parallel adapters can refine the feature representations from the original backbone when fine-tuning towards a new task. 
	
	Each adapter contains down-projection matrix $\textbf{W}_{down}\in \mathbb{R}^{n*m}$ to project the input to a lower-dimensional space, followed by a non-linear activation function $\sigma(\cdot)$, and an up-projection matrix $\textbf{W}_{up}\in \mathbb{R}^{m*n}$. Here, $n$ represents the dimension of the hidden layer, and $m$ serves as the adapter bottleneck dimension, a hyperparameter used in configuring the adapters. Denoting $\textbf{h}_{in}$ as the input to the adapter, the computation within the adapter module (with residual) can be summarized as follows:
	\begin{equation}
		\textbf{h}_{in} \leftarrow \textbf{h}_{in}+\sigma(\textbf{h}_{in} \textbf{W}_{down}) \textbf{W}_{up}.   
	\end{equation}
	
	Compared with previous side-tuning methods using lightweight transformer structures pruned from the backbone as the side network, our side-network design eliminates multi-head attention mechanisms and feed-forward networks. As a result, it further reduce amount of trainable parameters and increase the fine-tuning overhead.

	\textbf{Shortcut connection with quantized activation.} MobiLLM utilizes shortcut connections from the intermediate activations of the backbone model to the side network. The outputs of the frozen backbone’s final layer and the side network are combined and fed in the LLM head to predict. 
	
	MobiLLM allows the transformer layers of the pre-trained backbone to be grouped into $M$ blocks, with each block connected to a parallel adapter in the task-specific side network. Accordingly, there are $M$ shortcuts that feed intermediate activations from the backbone to the corresponding adapters in the side network. Let $\mathbf{a}_1, …, \mathbf{a}_m,…,\mathbf{a}_M$, where $a_m \in \mathbb{R}^{n\times d}$, denote the activation outputs with each consisting of $n$ tokens with a hidden dimension of $d$. To reduce communication overhead, MobiLLM applies low-precision quantization on these intermediate activations before feeding in the side-network. This process involves converting the original data format of activations (e.g., 32-bit or 16-bit floating-point) into a lower-bit data type. Typically, the input data is normalized by the absolute maximum value of its elements to fit within the range of the target data type. Taking FP4 quantization as an example, a 16-bit floating-point tensor is quantized into a 4-bit format with a range of $[0, 15]$ through
	
		\begin{equation}
			\mathbf{X}^{\mathrm{FP4}}=\operatorname{round}\left(\frac{15}{\operatorname{absmax}\left(\mathbf{X}^{\mathrm{FP} 16}\right)} \mathbf{X}^{\mathrm{FP} 16}\right),
		\end{equation}
	where $ \mathbf{X}^{\mathrm{FP}16} $ is the floating-point tensor, $ \mathbf{X}^{\mathrm{4bit} } $ is the quantized counterpart with $\mathrm{INT}4$ data type. Alternatively, by using NF4 data type, MobiLLM also allows for the following quantization process:    
	\begin{equation}
		\mathbf{X}^{\mathrm{NF}4} = \mathbf{T}\left(\frac{\mathbf{X}^{FP16}}{\text{absmax}(\mathbf{X}^{FP16})}\right),
	\end{equation}
	
	Here, the $ \mathbf{X}^{\mathrm{FP} 16} $ is first normalized using its maximum absolute value, and then the 4-bit quantized counterpart is obtained by mapping to a quantile table (denoted by $\mathbf{T}(\cdot)$). The quantile table $\mathbf{T}(\cdot)$ is well constructed to ensure that each quantization bin has an equal number of values assigned from the input tensor~\cite{dettmers2024qlora}. MobiLLM is potentially compatible with various quantization techniques and data formats, which may help mitigate information distortion by leveraging the distribution characteristics of activation values. We tested both FP4 and NF4 data types in our experiments and empirically demonstrated that NF4 achieves better performance (Section 5.4).  
	
	We note that it is not mandatory to partition the backbone into blocks with an identical number of transformer layers. A practical approach is to group the top layers (near the backbone’s output) more finely, while grouping the bottom layers more coarsely. The rationale is that the bottom layers of the pre-trained backbone often learn general low-level feature representations shared across downstream tasks, requiring minimal adjustment during fine-tuning. This eliminates the need for attaching tunable adapters to every transformer layer, further reducing communication overhead for uploading activations to the server. Moreover, this design enhances data privacy, as the activation outputs from each backbone block almost distort the input embeddings, making it difficult to infer the original input samples.

\subsection{MobiLLM Procedure}\label{sec:procedure}

This subsection details the training procedure of the proposed MobiLLM. We start with MobiLLM split learning within each iteration, and then present MobiLLM overlapping training strategy across iterations for better time efficiency.

\begin{figure}
	\centering
	\includegraphics[width=0.94\linewidth]{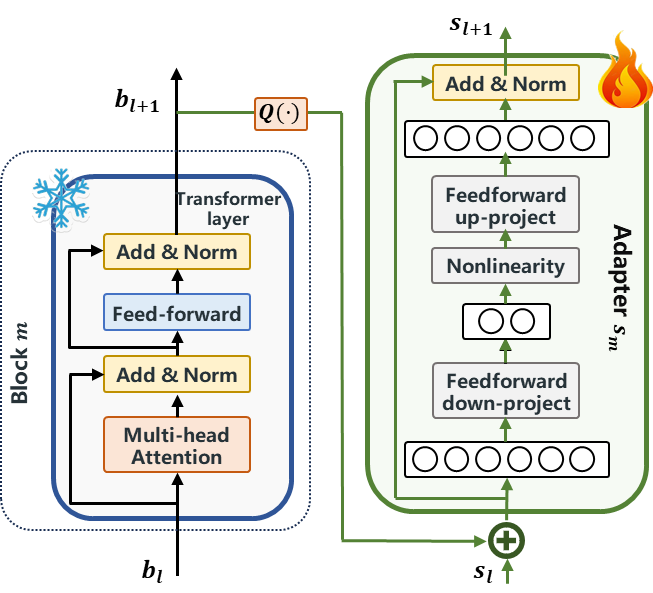}
	\caption{Schematic diagram of the quantized adapter side-tuning structure.}
	\label{fig:Fwcalculation}
\end{figure}

\textbf{MobiLLM split learning between the mobile device and the server in each iteration.} For each iteration, MobiLLM splits the training workloads between the mobile device and the server. Traditional split learning involves device-side forward propagation, server-side forward and backward propagation, and device-side backward propagation, requiring bidirectional exchanges of smashed data at the cut layer. Different from the traditional one, MobiLLM split learning simplifies this process by delegating backpropagation computations entirely to the server, thanks to its design separating the trainable side network from the frozen backbone network as shown in Fig.~\ref{fig:overview} and Fig.~\ref{fig:Fwcalculation}. Therefore, MobiLLM only requires a one-way transfer of intermediate activations (i.e., smashed data) from the mobile device to the server during the forward pass. Without loss of generality, we present the detailed training process of MobiLLM within an iteration as follows:

\begin{enumerate}
	\item \textit{Initialization (mobile device $\&$ server):} The mobile device retrieves a pre-trained LLM backbone that is suitable for its local computational and memory capabilities. Then, the server initializes a side-network, where the weights in the adapter modules were drawn from a zero-mean Gaussian with a well-selected standard deviation.  
	\item \textit{Local backbone forward propagation (mobile device):} The mobile device samples a mini-batch of data from its local dataset and performs forward propagation through the frozen backbone model, generating intermediate activation outputs corresponding to each transformer block.
	\item \textit{Activation transmission (mobile device $\rightarrow$ server):} The mobile device quantizes the intermediate activation outputs of the backbone model ($\overline{\mathbf{a}_1}, …, \overline{\mathbf{a}_m},…,\overline{\mathbf{a}_M}$) into a compressed format. The mobile device then transmits the quantized activations along with the corresponding labels and metadata (e.g., batch indices) to the server.
	\item \textit{Forward propagation (server):} Upon receiving the activations, the server integrates them with the input to its adapter modules, which serve as the side-network. The server performs forward propagation through the side-network, generating outputs needed for the training process. 
	\item \textit{Loss calculation (server):} The server computes the training loss by comparing the side-network's output with the ground-truth labels received from the mobile device. The loss metric depends on the specific downstream task (e.g., cross-entropy loss for classification or mean squared error for regression).
	\item \textit{Backward propagation and model updates (server):} The server performs backward propagation to update the side-network.
	\item The mobile device continues sampling new data batches and feeding them into the local backbone model. The above steps are repeated iteratively to tune the side-network until it converges.
\end{enumerate}

Figure~\ref{fig:Fwcalculation} illustrates the forward propagation process in the MobiLLM framework, using a GPT-style transformer architecture~\cite{brown2020language} as an example. For simplicity, key components within each transformer layer are denoted as follows: Multi-Head Self-Attention (MSA) is $f_{B_1}$, and Feed Forward Network (FFN) is $f_{B_3}$. The two Normalization (LN2) layers are denoted by $f_{B_2}$ and $f_{B_4}$, respectively. In the side-network, the down/up projections and non-linear layer in an adapter module are simplified as $f_{A_1}$, while Layer Normalization is represented by $f_{A_2}$. In the $l$-th transformer layer, the model generates two distinct outputs - one for the backbone and one for the side-network, which can be expressed as
\begin{equation}\label{Backfw}
	b_{l+1}=f_{B_4}(f_{B_3}(f_{B_2}(f_{B_1}(b_l)+b_l))+f_{B_2}(f_{B_1}(b_l)+b_l)),
\end{equation}
and
\begin{equation}\label{Sidefw}
	s_{l+1}=f_{A_2}(f_{A_1}(s_l+b_{l+1})+s_l+Q(b_{l+1})).
\end{equation}
Here, $Q(\cdot)$ denotes the quantization operator for the shortcut activation connections.

In our MobiLLM, the forward propagation calculations in Eq.~\ref{Backfw} and Eq.~\ref{Sidefw} are executed by the mobile device and server, respectively, and can be computed in parallel. Importantly, we observe the updates to the adapters are decoupled from the backpropagation of the backbone (i.e., Eq.~\ref{Backfw} operates independently of Eq.~\ref{Sidefw}), because the adapters are not embedded within the backbone layers. Besides, the adapters take the intermediate activations from the backbone as their inputs, which ensures that the side-network benefits from the pre-trained representations during training.

Next, we present the gradient calculation for the trainable parameters $\theta_l$ of the $l$-th adapter, where $\theta_l$ = $\{\theta_l^1,\theta_l^2,...,\theta_l^n\}$. We use $a_{l+1}$ to denote the concatenation $l$-th layer output of the backbone (i.e., $b_{l+1}$) and the side-network (i.e., $s_{l+1}$). In backpropagation with loss $L$, the gradient with respect to $\theta_l$ is:
\begin{equation}
	\begin{aligned}
		\frac{\partial L}{\partial \theta_l} &= \frac{\partial L}{\partial a_{l+1}} \frac{\partial a_{l+1}}{\partial f_A} \frac{\partial f_A}{\partial \theta_l} \\
		&= \frac{\partial L}{\partial s_{l+1}} \left( \frac{\partial s_{l+1}}{\partial f_l^1} \frac{\partial f_l^1}{\partial f_l^2} \frac{\partial f_l^2}{\partial f_l^3} \cdots \frac{\partial f_l^m}{\partial f_A} \right) \sum_{k=1}^n \frac{\partial f_A}{\partial \theta_l^k},	
	\end{aligned}
\end{equation}
where $f_l^m$ is the intermediate process from the output of the $l$-the layer to $f_A$. Most of existing PEFT methods only reduce the number of trainable parameters $\theta_l$, i.e., $\left| \sum_{k=1}^n \frac{\partial f_A}{\partial \theta_l^k} \right|$, resulting in a slight reduction in memory usage. In contrast, MobiLLM takes a step further. It simplifies the intermediate process $\frac{\partial a_{l+1}}{\partial f_A}$ and offloads the necessary gradient calculations to the memory-rich server along a parallel gradient highway. Therefore, MobiLLM saves considerable memory and computation time for the mobile device. Besides, MobiLLM effectively bypasses the inefficiencies of mobile accelerators in handling training-specific operations and ensures high utilization of available hardware resources on the mobile device. Since the server has much higher processing speeds than the mobile device, the server-side operations also speed up the LLM fine-tuning for resource-constrained mobile devices.

\textbf{MobiLLM overlapping device-side and server-side trainings across iterations.} MobiLLM diverges from the conventional one-forward-one-backward interleaved updating rule by enabling uninterrupted forward pass computations on the mobile device. In parallel with intermediate activation transmission and side-network training, the mobile device overlaps the process by continuously feeding data batches into the backbone for successive training iterations. This is grounded in the fact that the device-side backbone model remains fixed throughout the fine-tuning process, eliminating the need to wait for parameter updates. As a result, multi-batch parallelism is achieved without introducing model staleness, which is a common drawback in other overlapping training methods. Besides, the server-side side-network training, powered by high-performance computation, typically outpaces the device’s execution and transmission speeds by several times. This streamlined workflow guarantees the timely reception of up-to-date intermediate outputs by the server, eliminating unnecessary waiting times and accelerating collaborative fine-tuning.

\subsection{Advantages of MobiLLM}
In this subsection, we first analyze how MobiLLM helps to save memory of LLM fine-tuning on the mobile device, and then discuss MobiLLM's other merits.

\textbf{Analysis of memory saving.} MobiLLM restricts device-side operations to forward propagation only, drastically reducing the mobile device's memory usage. Specifically, MobiLLM reduces two main contributors of the memory footprint for the mobile device during the fine-tuning process, i.e., intermediate activations, optimizer states and model states, which are analyzed as follows.

1) Intermediate activations: During forward propagation, only the activations of a few selected layers are temporarily stored on the device and can be discarded after transmission to the server. Consider an $L$-layer LLM, where the activation size of each layer is denoted by $A=S \cdot B \cdot H$. Here, $S$, $B$, and $H$ represent the sequence length, batch size, and hidden-layer dimension, respectively. Traditional fine-tuning methods require storing activations for all layers on the mobile device, which consumes approximately $L \cdot A$ of memory. In contrast, MobiLLM requires only $\gamma \cdot A$, where $\gamma$ ($\gamma \ll L$) is the number of selected layers for temporary storage. This reduces memory usage for activations by approximately $L/\gamma$-fold.

2) Optimizer State: Classical optimization algorithms like Adam require storing additional states (e.g., momentum and second-order statistics), typically doubling or tripling the memory needed for trainable model parameters. In MobiLLM, these computations and states are fully offloaded to the server, avoiding that memory consumption on the mobile device.


Those optimizations in MobiLLM design collectively reduce the memory footprint, enabling mobile devices even with limited memory capacities to fine-tune LLMs efficiently.

\textbf{Advantages beyond memory efficiency.} MobiLLM has other merits that enhance its practicality and performance in resource-constrained settings, which are listed as follows.

1) Enhanced system reliability. Rather than relying on flimsy multi-hop pipeline collaboration among multiple mobile devices~\cite{borzunov2024distributed,wang2024federated}, MobiLLM only employs a single server with sufficient computational resources for help. Such a simple device-server design aligns with existing mobile edge network architecture, which has mature physical and network layer protocols to support terminal-server collaboration with high speed transmissions. For instance, 5G/6G base stations often feature high-performance computational units to provide nearby services for mobile users, and smart hubs in home IoT networks can manage workloads for connected devices via high-speed Wi-Fi links. Thus, the setup of MobiLLM is inherently more reliable and practical. Besides, MobiLLM protects local data privacy by retaining all data samples on the mobile device throughout the LLM fine-tuning process.

2) Versatility besides fine-tuning. Unlike pipeline-based methods that partition LLM across devices, MobiLLM allows each mobile device to keep a full and lightweight LLM. That enables mobile devices to independently provide local inference services besides on-device LLM fine-tuning. In other words, the mobile device may respond to on-demand and latency-sensitive inference requests during a continuous fine-tuning process via local resource sharing, i.e., simultaneously serving LLM fine-tuning and inference. Besides, the device can periodically update its model by fetching and merging task-specific adapters trained on the server, and the server can host multiple side-networks for different downstream tasks. That empowers a mobile device to switch between LLM fine-tuning tasks by simply altering the side-networks required at the server side. This flexibility enhances mobile devices’ utility both during and beyond LLM fine-tuning process.

\section{Experimental Setup}
\subsection{MobiLLM Implementation}

The proposed MobiLLM system is deployed on a testbed consisting of a server and a mobile device. The server is equipped with a NVIDIA A100 GPU (6912-core Ampere GPU with 40GB memory). On the mobile device side, we consider two classical types of mobile devices: (1) NVIDIA Jetson Xavier NX with 6-core NVIDIA Carmel ARM CPU, 384-core NVIDIA Volta GPU, and 8GB RAM. (2) Huawei Matebook laptop with Intel 13th Gen Core i5-13500H and 16GB RAM, which is a CPU-only device. The communication between the mobile device and the server is based on Wi-Fi 5 connections, which uses the WebSocket communication protocol.

In particular, LLMs training on Jetson Xavier NX contributes the major experimental results in this work. Since Xavier's GPU and CPU share 8GB RAM, its maximum available GPU memory for training is 4.6G. The Matebook laptop is employed to test the training acceleration performance of those CPU-only mobile devices (Sec.~\ref{subsec:adv}). 

\subsection{Models, Datasets and Parameters}

We exploit the OPT-350M and OPT-1.3B, two popular decoder-only LLMs from the OPT series, to evaluate MobiLLM's performance. The pre-trained weights of the above models are from Huggingface~\cite{wolf2019huggingface}.

We use natural language understanding (NLU) tasks to compare MobiLLM and several baselines on natural language understanding (NLU) tasks. We use the GLUE ~\cite{wang2018glue} benchmark, which consists of seven classification and one regression task. The benchmark evaluate models on multiple diverse tasks over linguistic acceptability (CoLA ~\cite{warstadt2019neural}), sentiment analysis (SST-2~\cite{socher2013recursive}), similarity and paraphrase (MRPC~\cite{dolan2005automatically}, QQP~\cite{iyer2017first}, STS-B~\cite{cer2017semeval}) and natural language inference (MNLI~\cite{williams2017broad}, QNLI~\cite{rajpurkar2016squad}, RTE~\cite{socher2013recursive}). We report accuracy on MNLI, QQP, QNLI, SST-2, MRPC, and RTE, Pearson correlation coefficients on SST-B, and Mathews correlation coefficients~\cite{matthews1975comparison} on CoLA. 

Unless otherwise specified, MobiLLM and all baseline methods use the same hyperparameter setup: FP16 (float16) training precision, 20 local training epochs, a mini-batch size of 16, a learning rate of 5e-4, and a maximum sequence length of 256. The network bandwidth is set to 60 Mbps.

\begin{table*}[]
		\captionsetup{justification=centering}
		\caption{Experiment results on GLUE benchmark. \\ Model: OPT-350M (Batch size=16, Sequence length=256)}
		\label{table:comparisonSOTA350}
		\small
		\setlength{\tabcolsep}{4.5pt}
		\vspace{-2mm}
		\begin{tabular}{c|c|c|cc|ccccccccc}
			\hline
			\multirow{2}{*}{Method} &
			\multirow{2}{*}{\begin{tabular}[c]{@{}c@{}}Hidden \\ Size/ r\end{tabular}} &
			\multirow{2}{*}{\begin{tabular}[c]{@{}c@{}}Trainable \\ Parameters (\%)\end{tabular}} &
			\multicolumn{2}{c|}{Memory Usage (GB)} &
			\multicolumn{9}{c}{Accuracy} \\ \cline{4-14} 
			&
			&
			&
			\multicolumn{1}{c|}{Training} &
			Inference &
			\multicolumn{1}{c|}{SST-2} &
			\multicolumn{1}{c|}{QNLI} &
			\multicolumn{1}{c|}{QQP} &
			\multicolumn{1}{c|}{MRPC} &
			\multicolumn{1}{c|}{CoLA} &
			\multicolumn{1}{c|}{MNLI} &
			\multicolumn{1}{c|}{RTE} &
			\multicolumn{1}{c|}{STS-B} &
			Avg \\ \hline
			Full-FT &
			---  &
			100 &
			\multicolumn{1}{c|}{7.910} &
			1.580 &
			\multicolumn{1}{c|}{92.9} &
			\multicolumn{1}{c|}{\textbf{84.2}} &
			\multicolumn{1}{c|}{\textbf{85.9}} &
			\multicolumn{1}{c|}{\textbf{83}} &
			\multicolumn{1}{c|}{\textbf{61.9}} &
			\multicolumn{1}{c|}{76.4} &
			\multicolumn{1}{c|}{\textbf{76}} &
			\multicolumn{1}{c|}{85.6} &
			\multicolumn{1}{l}{\textbf{80.7}} \\ \hline
			\multirow{2}{*}{LoRA} &
			64 &
			1.86 &
			\multicolumn{1}{c|}{6.700} &
			1.580 &
			\multicolumn{1}{c|}{92.7} &
			\multicolumn{1}{c|}{84.1} &
			\multicolumn{1}{c|}{85.1} &
			\multicolumn{1}{c|}{82.2} &
			\multicolumn{1}{c|}{61.1} &
			\multicolumn{1}{c|}{\textbf{76.5}} &
			\multicolumn{1}{c|}{75.8} &
			\multicolumn{1}{c|}{85.9} &
			80.4 \\ \cline{2-14} 
			&
			16 &
			0.47 &
			\multicolumn{1}{c|}{6.655} &
			1.580 &
			\multicolumn{1}{c|}{92.2} &
			\multicolumn{1}{c|}{83.7} &
			\multicolumn{1}{c|}{84.3} &
			\multicolumn{1}{c|}{81.7} &
			\multicolumn{1}{c|}{60.4} &
			\multicolumn{1}{c|}{76.1} &
			\multicolumn{1}{c|}{75.4} &
			\multicolumn{1}{c|}{85.6} &
			79.9 \\ \hline
			BitFit &
			--- &
			0.08 &
			\multicolumn{1}{c|}{6.081} &
			1.580 &
			\multicolumn{1}{c|}{\textbf{93.8}} &
			\multicolumn{1}{c|}{\textbf{84.2}} &
			\multicolumn{1}{c|}{84.6} &
			\multicolumn{1}{c|}{82.2} &
			\multicolumn{1}{c|}{61.4} &
			\multicolumn{1}{c|}{76.3} &
			\multicolumn{1}{c|}{\textbf{76}} &
			\multicolumn{1}{c|}{\textbf{86.2}} &
			80.6 \\ \hline
			\multirow{2}{*}{LST} &
			16 &
			0.84 &
			\multicolumn{1}{c|}{4.932} &
			1.624 &
			\multicolumn{1}{c|}{91.9} &
			\multicolumn{1}{c|}{84} &
			\multicolumn{1}{c|}{84.5} &
			\multicolumn{1}{c|}{82.3} &
			\multicolumn{1}{c|}{60.2} &
			\multicolumn{1}{c|}{75.8} &
			\multicolumn{1}{c|}{73.9} &
			\multicolumn{1}{c|}{84.9} &
			79.7 \\ \cline{2-14} 
			&
			64 &
			0.15 &
			\multicolumn{1}{c|}{4.787} &
			1.588 &
			\multicolumn{1}{c|}{91.4} &
			\multicolumn{1}{c|}{83.8} &
			\multicolumn{1}{c|}{84.3} &
			\multicolumn{1}{c|}{81.9} &
			\multicolumn{1}{c|}{60.2} &
			\multicolumn{1}{c|}{75.6} &
			\multicolumn{1}{c|}{73.2} &
			\multicolumn{1}{c|}{84.3} &
			79.3 \\ \hline
			\multirow{2}{*}{MobiLLM-L} &
			64 &
			1.12 &
			\multicolumn{1}{c|}{\textbf{2.452}} &
			1.623 &
			\multicolumn{1}{c|}{90.2} &
			\multicolumn{1}{c|}{83.7} &
			\multicolumn{1}{c|}{84.2} &
			\multicolumn{1}{c|}{81.8} &
			\multicolumn{1}{c|}{59.7} &
			\multicolumn{1}{c|}{75.2} &
			\multicolumn{1}{c|}{73.4} &
			\multicolumn{1}{c|}{84.2} &
			79.1 \\ \cline{2-14} 
			&
			16 &
			0.42 &
			\multicolumn{1}{c|}{\textbf{2.421}} &
			1.605 &
			\multicolumn{1}{c|}{90} &
			\multicolumn{1}{c|}{83.4} &
			\multicolumn{1}{c|}{84} &
			\multicolumn{1}{c|}{81.6} &
			\multicolumn{1}{c|}{59.3} &
			\multicolumn{1}{c|}{75.1} &
			\multicolumn{1}{c|}{72.9} &
			\multicolumn{1}{c|}{83.7} &
			78.8 \\ \hline
			\multirow{2}{*}{MobiLLM} &
			64 &
			0 &
			\multicolumn{1}{c|}{\textbf{1.635}} &
			1.623 &
			\multicolumn{1}{c|}{89.9} &
			\multicolumn{1}{c|}{83.5} &
			\multicolumn{1}{c|}{83.9} &
			\multicolumn{1}{c|}{81.7} &
			\multicolumn{1}{c|}{59.4} &
			\multicolumn{1}{c|}{75.1} &
			\multicolumn{1}{c|}{73} &
			\multicolumn{1}{c|}{84} &
			78.8 \\ \cline{2-14} 
			&
			16 &
			0 &
			\multicolumn{1}{c|}{\textbf{1.621}} &
			1.605 &
			\multicolumn{1}{c|}{89.8} &
			\multicolumn{1}{c|}{83.1} &
			\multicolumn{1}{c|}{83.5} &
			\multicolumn{1}{c|}{81.6} &
			\multicolumn{1}{c|}{59.3} &
			\multicolumn{1}{c|}{74.7} &
			\multicolumn{1}{c|}{72.7} &
			\multicolumn{1}{c|}{83.5} &
			78.5 \\ \hline
		\end{tabular}
	\end{table*}

	\begin{table*}[]
		\captionsetup{justification=centering}
		\caption{Experiment results on GLUE benchmark. \\ Model: OPT-1.3B (Batch size=16, Sequence length=256)}
		\label{table:comparisonSOTA}
		\small
		\setlength{\tabcolsep}{4.5pt}
		\vspace{-2mm}
		\begin{tabular}{c|c|c|cc|ccccccccc}
			\hline
			\multirow{2}{*}{Method} &
			\multirow{2}{*}{\begin{tabular}[c]{@{}c@{}}Hidden \\ Size/ r\end{tabular}} &
			\multirow{2}{*}{\begin{tabular}[c]{@{}c@{}}Trainable \\ Parameters (\%)\end{tabular}} &
			\multicolumn{2}{c|}{Memory Usage (GB)} &
			\multicolumn{9}{c}{Accuracy} \\ \cline{4-14} 
			&
			&
			&
			\multicolumn{1}{c|}{Training} &
			Inference &
			\multicolumn{1}{c|}{SST-2} &
			\multicolumn{1}{c|}{QNLI} &
			\multicolumn{1}{c|}{QQP} &
			\multicolumn{1}{c|}{MRPC} &
			\multicolumn{1}{c|}{CoLA} &
			\multicolumn{1}{c|}{MNLI} &
			\multicolumn{1}{c|}{RTE} &
			\multicolumn{1}{c|}{STS-B} &
			Avg \\ \hline
			Full-FT &
			---  &
			100 &
			\multicolumn{1}{c|}{20.895} &
			4.403 &
			\multicolumn{1}{c|}{\textbf{95.9}} &
			\multicolumn{1}{c|}{85} &
			\multicolumn{1}{c|}{\textbf{86.9}} &
			\multicolumn{1}{c|}{\textbf{84}} &
			\multicolumn{1}{c|}{\textbf{63.9}} &
			\multicolumn{1}{c|}{81} &
			\multicolumn{1}{c|}{\textbf{83}} &
			\multicolumn{1}{c|}{\textbf{89.6}} &
			\multicolumn{1}{l}{\textbf{83.6}} \\ \hline
			\multirow{2}{*}{LoRA} &
			64 &
			0.95 &
			\multicolumn{1}{c|}{14.569} &
			4.403 &
			\multicolumn{1}{c|}{94.4} &
			\multicolumn{1}{c|}{84.9} &
			\multicolumn{1}{c|}{86.6} &
			\multicolumn{1}{c|}{83.3} &
			\multicolumn{1}{c|}{62.5} &
			\multicolumn{1}{c|}{\textbf{81.1}} &
			\multicolumn{1}{c|}{82.3} &
			\multicolumn{1}{c|}{89.2} &
			83.0 \\ \cline{2-14} 
			&
			16 &
			0.24 &
			\multicolumn{1}{c|}{14.497} &
			4.403 &
			\multicolumn{1}{c|}{94.2} &
			\multicolumn{1}{c|}{84.6} &
			\multicolumn{1}{c|}{86.1} &
			\multicolumn{1}{c|}{82.9} &
			\multicolumn{1}{c|}{61.9} &
			\multicolumn{1}{c|}{79.7} &
			\multicolumn{1}{c|}{82.1} &
			\multicolumn{1}{c|}{88.7} &
			82.5 \\ \hline
			BitFit &
			---  &
			0.04 &
			\multicolumn{1}{c|}{13.371} &
			4.403 &
			\multicolumn{1}{c|}{95.4} &
			\multicolumn{1}{c|}{85.5} &
			\multicolumn{1}{c|}{86.4} &
			\multicolumn{1}{c|}{83.3} &
			\multicolumn{1}{c|}{62.2} &
			\multicolumn{1}{c|}{79.9} &
			\multicolumn{1}{c|}{81.7} &
			\multicolumn{1}{c|}{88.8} &
			82.9 \\ \hline
			\multirow{2}{*}{LST} &
			16 &
			0.85 &
			\multicolumn{1}{c|}{11.122} &
			4.533 &
			\multicolumn{1}{c|}{94.5} &
			\multicolumn{1}{c|}{\textbf{85.8}} &
			\multicolumn{1}{c|}{86.4} &
			\multicolumn{1}{c|}{83.3} &
			\multicolumn{1}{c|}{60.8} &
			\multicolumn{1}{c|}{77.6} &
			\multicolumn{1}{c|}{81.6} &
			\multicolumn{1}{c|}{88.4} &
			82.3 \\ \cline{2-14} 
			&
			64 &
			0.15 &
			\multicolumn{1}{c|}{10.804} &
			4.427 &
			\multicolumn{1}{c|}{94.4} &
			\multicolumn{1}{c|}{85.6} &
			\multicolumn{1}{c|}{86.2} &
			\multicolumn{1}{c|}{83.1} &
			\multicolumn{1}{c|}{60.3} &
			\multicolumn{1}{c|}{77.4} &
			\multicolumn{1}{c|}{81.2} &
			\multicolumn{1}{c|}{88.1} &
			82.0 \\ \hline
			\multirow{2}{*}{MobiLLM-L} &
			64 &
			0.49 &
			\multicolumn{1}{c|}{\textbf{6.132}} &
			4.472 &
			\multicolumn{1}{c|}{94.1} &
			\multicolumn{1}{c|}{84.7} &
			\multicolumn{1}{c|}{85.8} &
			\multicolumn{1}{c|}{81.7} &
			\multicolumn{1}{c|}{59.9} &
			\multicolumn{1}{c|}{76.9} &
			\multicolumn{1}{c|}{80.7} &
			\multicolumn{1}{c|}{87.9} &
			81.5 \\ \cline{2-14} 
			&
			16 &
			0.13 &
			\multicolumn{1}{c|}{\textbf{6.088}} &
			4.461 &
			\multicolumn{1}{c|}{94.1} &
			\multicolumn{1}{c|}{84.3} &
			\multicolumn{1}{c|}{85.6} &
			\multicolumn{1}{c|}{81.2} &
			\multicolumn{1}{c|}{59.4} &
			\multicolumn{1}{c|}{76.7} &
			\multicolumn{1}{c|}{80.4} &
			\multicolumn{1}{c|}{87.5} &
			81.2 \\ \hline
			\multirow{2}{*}{MobiLLM} &
			64 &
			0 &
			\multicolumn{1}{c|}{\textbf{4.495}} &
			4.472 &
			\multicolumn{1}{c|}{94} &
			\multicolumn{1}{c|}{84.5} &
			\multicolumn{1}{c|}{85.4} &
			\multicolumn{1}{c|}{81.7} &
			\multicolumn{1}{c|}{59.7} &
			\multicolumn{1}{c|}{76.8} &
			\multicolumn{1}{c|}{80.6} &
			\multicolumn{1}{c|}{87.3} &
			81.3 \\ \cline{2-14} 
			&
			16 &
			0 &
			\multicolumn{1}{c|}{\textbf{4.487}} &
			4.461 &
			\multicolumn{1}{c|}{93.8} &
			\multicolumn{1}{c|}{84.1} &
			\multicolumn{1}{c|}{85.3} &
			\multicolumn{1}{c|}{81.4} &
			\multicolumn{1}{c|}{59.4} &
			\multicolumn{1}{c|}{76.6} &
			\multicolumn{1}{c|}{80.2} &
			\multicolumn{1}{c|}{87.3} &
			81.0 \\ \hline
		\end{tabular}
	\end{table*}
\subsection{Baselines for Comparison}

We compare MobiLLM with the following baselines under different LLM tasks. 

\noindent\textbf{Full-FT}: Fine-tune the whole LLM.  \\
\textbf{LoRA~\cite{hulora}}: Insert trainable low-rank matrices into the pre-trained LLM while keeping the backbone frozen, where the hidden size indicates the low-rank dimension.   \\
\textbf{BitFit~\cite{zaken2021bitfit}}: Fine-tune only the bias terms of the pre-trained LLM while keeping all other weights frozen.\\
\textbf{LST~\cite{sung2022lst}}: Use lightweight transformer structures pruned from the backbone as the side-network. Here, we use $r$ to indicate the side-network reduction factor, where a larger $r$ results in a smaller size of side-network. \\
\textbf{MobiLLM-L} (local): Train both the backbone network and the side-network locally on the mobile device.

\section{Evaluation Results \& Analysis}

\subsection{End-to-end Performance}

\textbf{Performance of different baselines on GLUE Benchmark.} We first study the performance of MobiLLM on the OPT-350M model. Table~\ref{table:comparisonSOTA350} summarizes the experimental results of MobiLLM and other baselines on GLUE benchmark under the default setting. Overall, MobiLLM achieves the lowest memory footprint among all methods while maintaining comparable accuracy. Particularly, MobiLLM outperforms Full-FT by a significant reduction of 79.5\% in memory usage, while only introducing a 2\% accuracy drop. Against other PEFT baselines, i.e., LoRA and BitFit, MobiLLM is much better than them by reducing averaged 75\% memory requirements. Using a local deployment without server assistance, MobiLLM-L reduces memory usage by 2.48 GB compared to LST, thanks to its lightweight adapter-based side-network design. With server-assisted side-tuning, MobiLLM further reduces memory usage to 1.621 GB. The significant improvement over all the baselines is mainly attributed to MobiLLM’s design, which offloads dominant computational and memory burdens to the high-performance server. It then allows the mobile device to handle only the forward pass, in contrast to other baselines that all require the backward pass on the mobile device and can easily result in out-of-memory errors. Moreover, MobiLLM enables the mobile device to fine-tune LLM with a memory cost close to that of on-device inference, as verified in Table~\ref{table:comparisonSOTA350}. That makes on-device LLM fine-tuning practical, especially for memory-constrained mobile devices.

\textbf{MobiLLM makes the billion-sized LLM model fine-tuning affordable to the mobile device.} We further evaluate MobiLLM using a billion-sized model, OPT-1.3B, and report the results in Table~\ref{table:comparisonSOTA}. LoRA, BitFit and LST have excessively huge memory footprints exceeding 10 GB, rendering them infeasible for most mainstream mobile devices. In contrast, our MobiLLM effectively supports fine-tuning such sizable models on a single mobile device. For instance, MobiLLM requires only 4.487 GB of memory on the mobile device to run the fine-tuning task smoothly on Xavier with 4.6 GB DRAM. Those results demonstrate that even resource-constrained mobile devices can reliably fine-tune such billion-sized LLMs without the risk of memory overflow.

\subsection{Advantages in Memory Efficiency}

\begin{figure}
    \centering
    \includegraphics[width=0.95\linewidth]{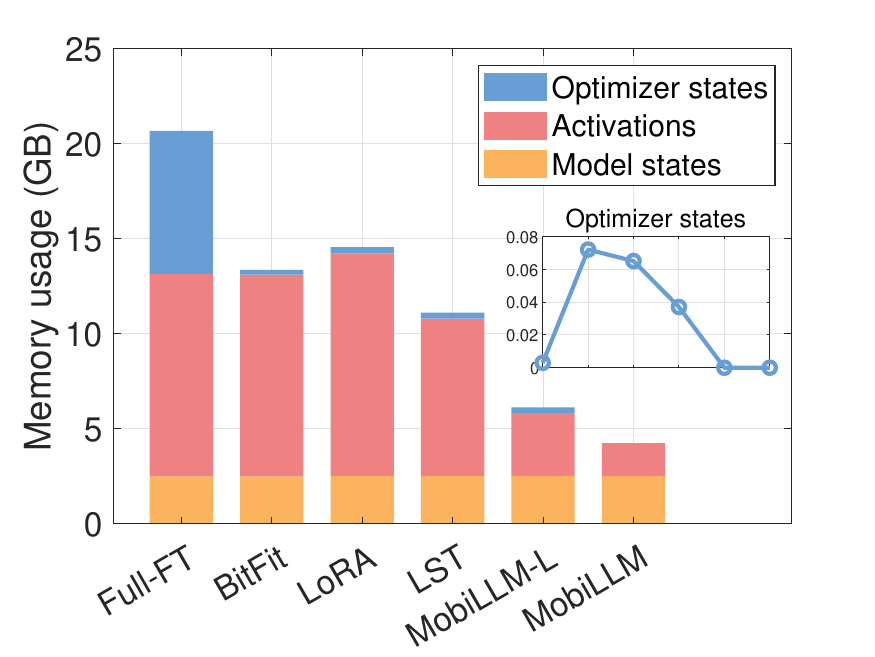}
    \caption{The breakdown of memory footprint. }
    \label{fig:memorybreak}
\end{figure}

\textbf{MobiLLM excels in memory saving.} Figure~\ref{fig:memorybreak} reports the breakdown of memory footprint for different methods when fine-tuning OPT-1.3B. Both LoRA and BitFit significantly reduce memory consumption by decreasing the number of trainable parameters, thereby remarkably lowering the memory required for optimizer states, as shown in the zoomed-in view in Fig.~\ref{fig:memorybreak}. MobiLLM and LST further optimize the memory usage by establishing backpropagation highways for side-tuning, which reduces the need to store intermediate activations for the backward pass. As a result, it achieves a total memory reduction of 2\textasciitilde 4 GB. MobiLLM also achieves an additional reduction of 6.4\textasciitilde6.7 GB memory compared to LST, owing to two key factors: 1) MobiLLM offloads the intermediate activation memory burden of the side-network to the server. 2) MobiLLM eliminates the need for storing the optimizer states at the mobile device since it makes the mobile device only execute forward propagation.
	
\begin{figure*} \centering 
		\subfigure[Impact of batch size (Model: OPT-1.3B)\label{mem-batch}]
		{\includegraphics[width=5.6cm, height=4.6cm]{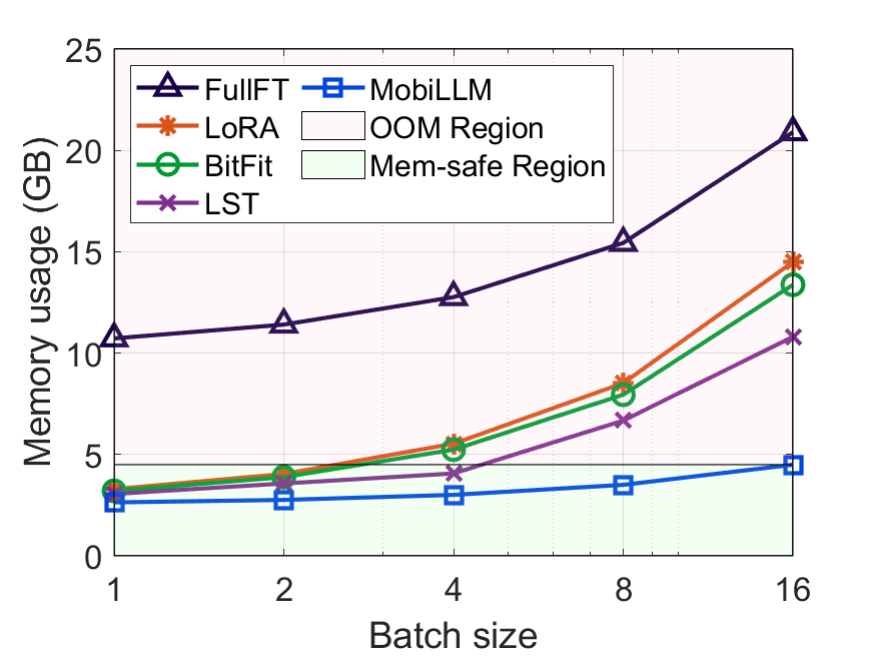}}
		\subfigure[Impact of sequence length \label{mem-seq}]
		{\includegraphics[width=5.6cm, height=4.6cm]{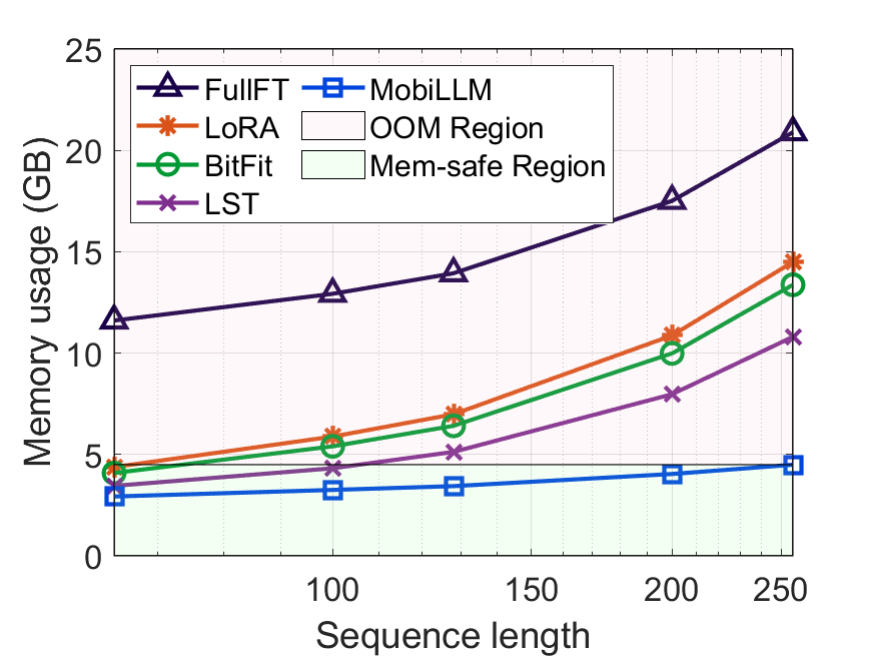}}
		\subfigure[Impact of batch size\label{mem-model}]
		{\includegraphics[width=5.6cm, height=4.6cm]{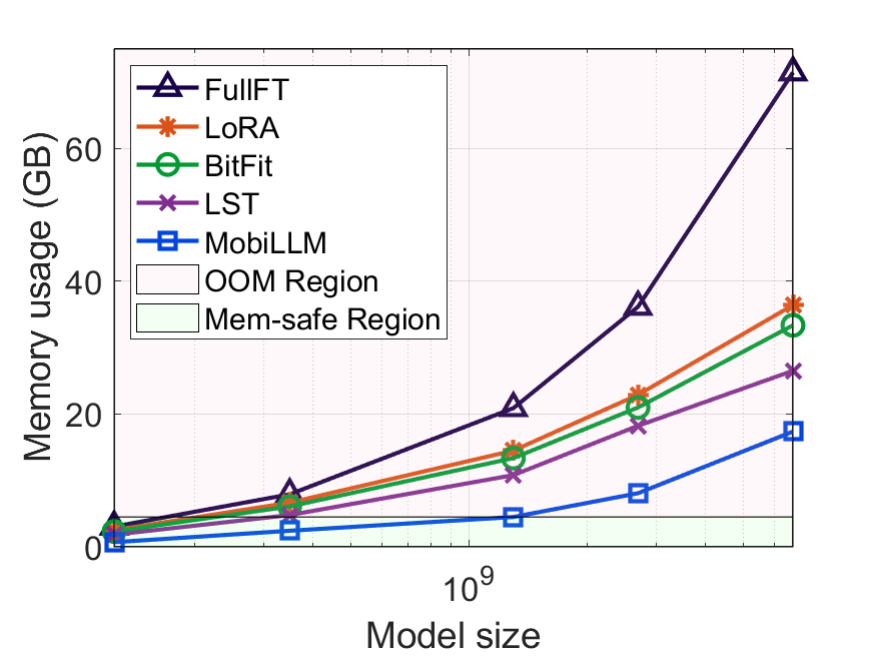}}
		\caption{\centering Performance on various training configurations. (Green area indicates the memory-safe region for Xaiver with peaking 4.6 GB GPU RAM)}   \label{Fig:MemEffect}
	\end{figure*}

\textbf{MobiLLM achieves near-stable memory usage across diverse training configurations.} Figure~\ref{mem-batch}-\ref{mem-seq} further explore how batch size, sequence length, and model size affect the memory usage. Figure~\ref{mem-batch} shows that while memory usage increases with batch size for all methods, MobiLLM consistently maintains the lowest memory footprint among all. Besides, MobiLLM exhibits slower memory growth compared to other baselines as the batch size increases. This is because, from a memory usage perspective, the batch size primarily affects the amount of intermediate activations that need to be stored for backpropagation, whereas our MobiLLM frees the mobile device from performing backpropagation entirely. Figure~\ref{mem-seq} examines memory usage for varying sequence lengths. Similar to the impact of batch size, LST and MobiLLM alleviate the growth rate of memory footprint of intermediate activations. Notably, MobiLLM requires only 41\% of the device-side memory usage of LST, when the sequence length is set to 256. Figure~\ref{mem-model} further evaluates the impact of different model sizes. Results indicate that MobiLLM consistently outperforms baselines, with its memory advantage widening as model size grows. These results verify that MobiLLM makes the mobile device's memory usage insensitive to training configurations such as batch size and sequence length, having a near-consistent memory cost. This alleviates the common ``occasional success or failure'' issue for fine-tuning specific models under varying settings. It further helps to prevent memory overflows on mobile devices even when processing larger batch sizes and longer sequences, ensuring reliable fine-tuning performance.

\subsection{Advantages in Fine-tuning Acceleration}
\label{subsec:adv}
\begin{figure*} \centering 
	\subfigure[SST-2@OPT-350M\label{sst350X}]
	{\includegraphics[width=4.35cm,height=3.6cm]{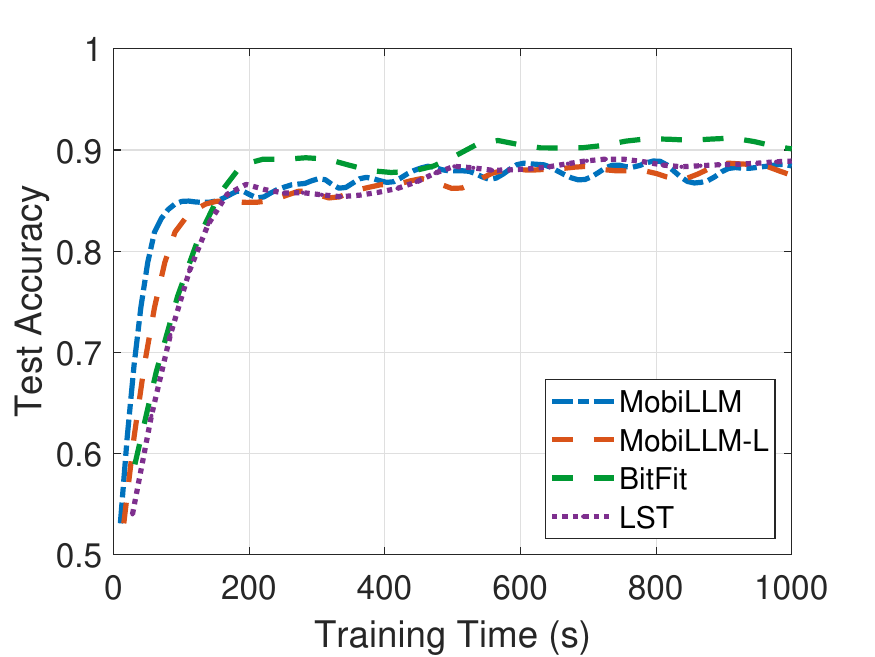}}
	\subfigure[SST-2@OPT-1.3B\label{sst13X}]
	{\includegraphics[width=4.35cm,height=3.6cm]{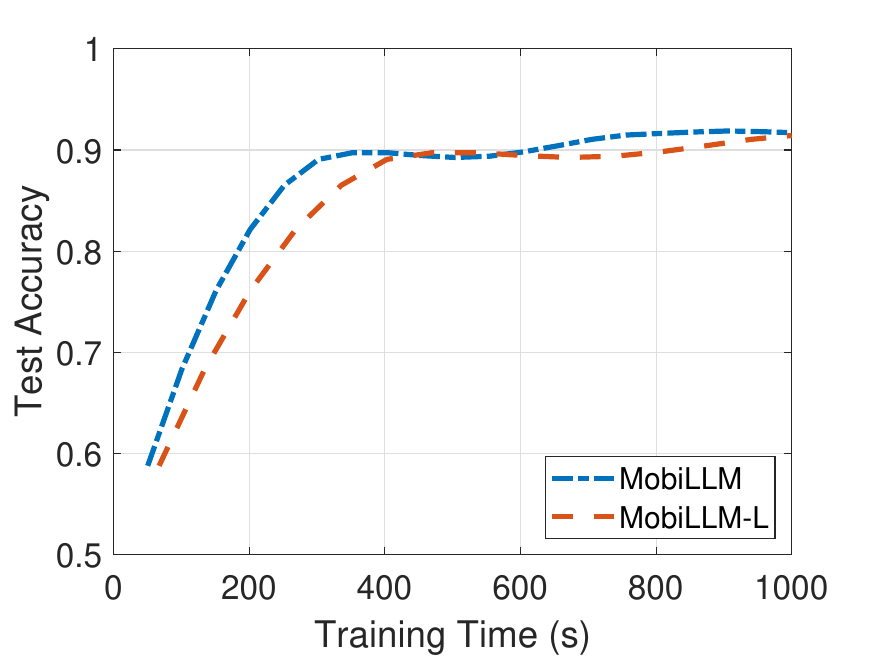}}
	\subfigure[MRPC@OPT-350M\label{mrpc350X}]
	{\includegraphics[width=4.35cm,height=3.6cm]{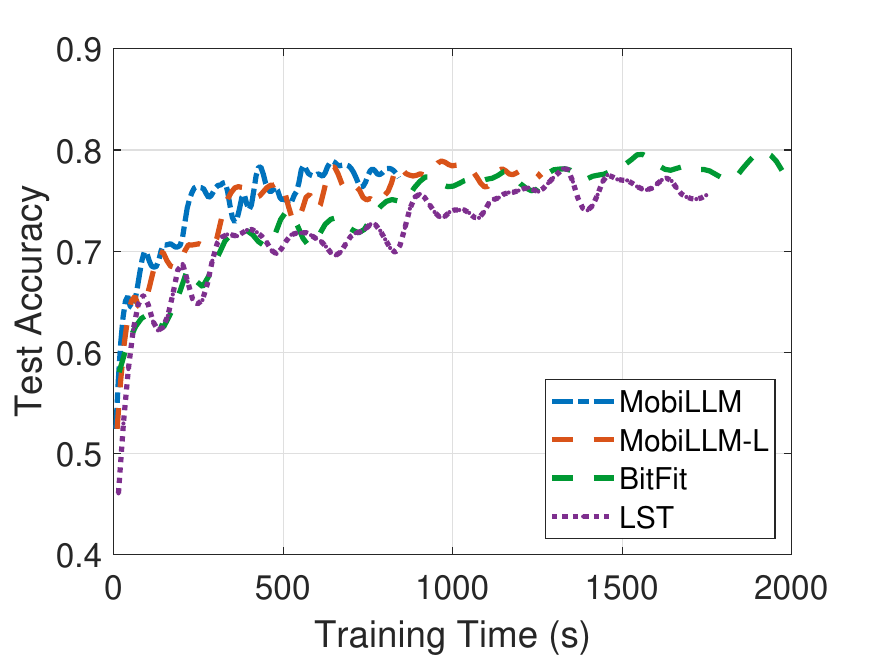}}
	\subfigure[MRPC@OPT-1.3B\label{mrpc13X}]
	{\includegraphics[width=4.35cm,height=3.6cm]{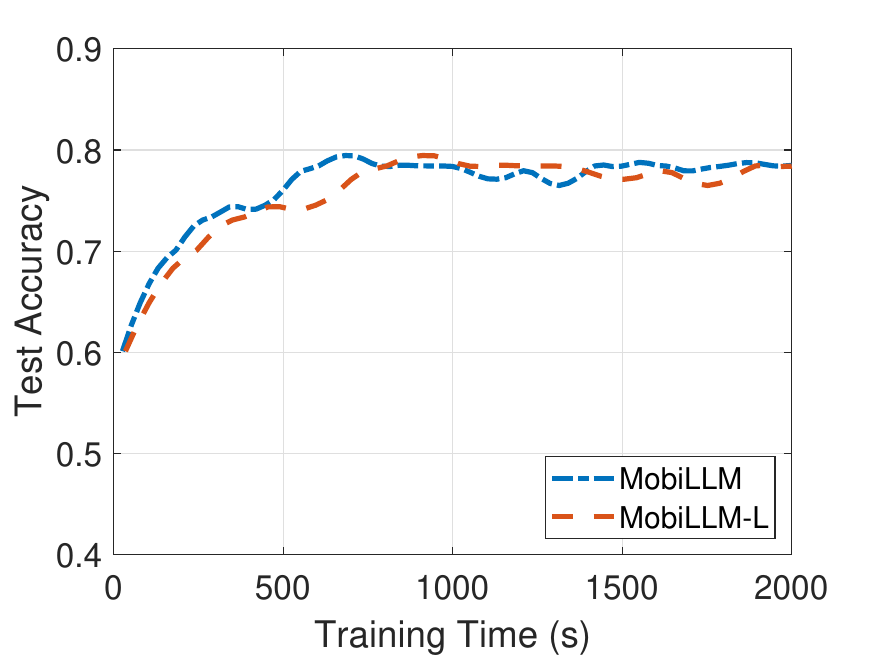}}
	\caption{Convergence performance on various models and tasks (On Xavier). \label{Fig:ConvOPTXavier}}
\end{figure*}

\begin{figure*} \centering 
	\subfigure[SST-2@OPT-350M\label{sst350L}]
	{\includegraphics[width=4.35cm,height=3.6cm]{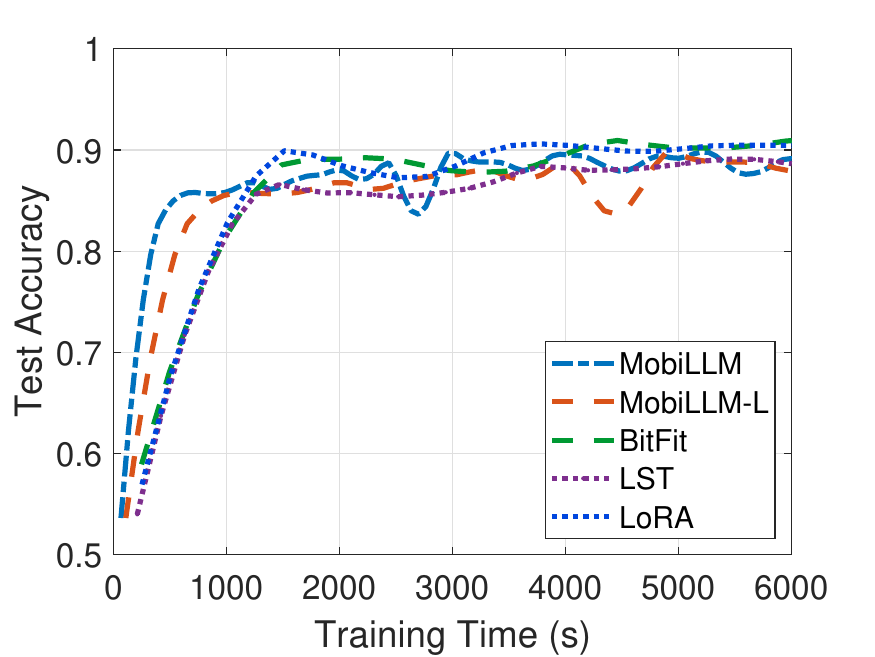}}
	\subfigure[SST-2@OPT-1.3B\label{sst13L}]
	{\includegraphics[width=4.35cm,height=3.6cm]{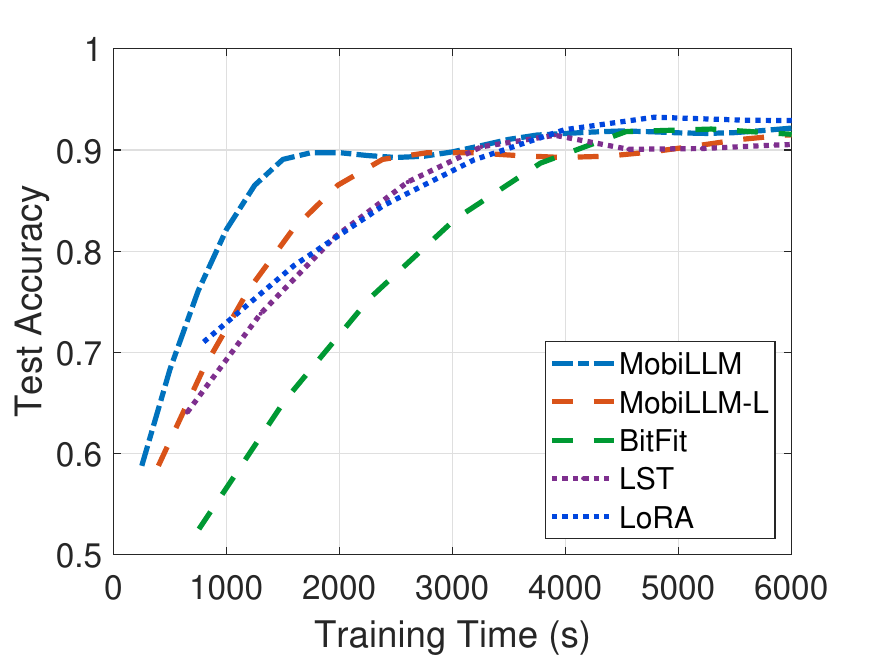}}
	\subfigure[MRPC@OPT-350M\label{mrpc350L}]
	{\includegraphics[width=4.35cm,height=3.6cm]{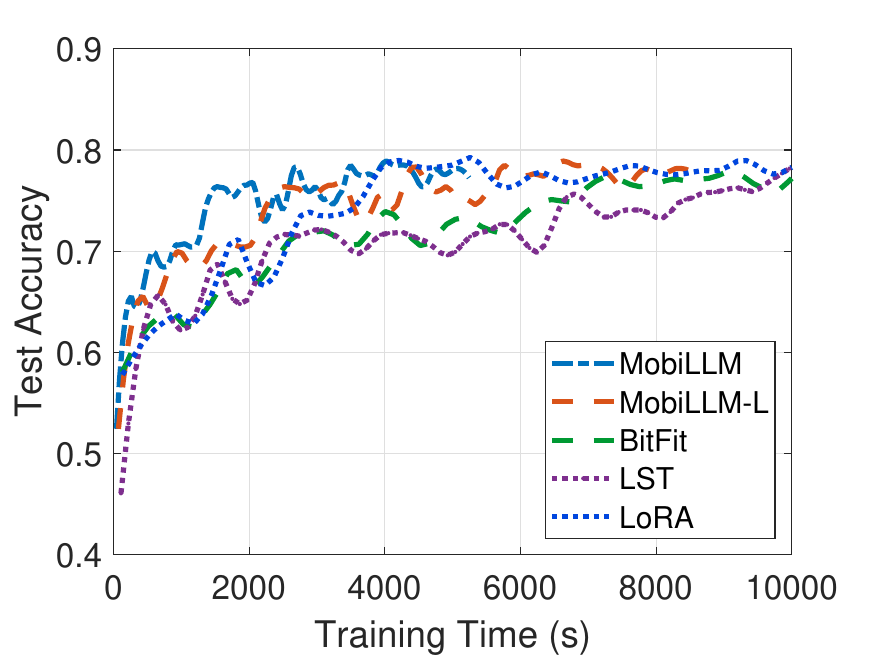}}
	\subfigure[MRPC@OPT-1.3B\label{mrpc13L}]
	{\includegraphics[width=4.35cm,height=3.6cm]{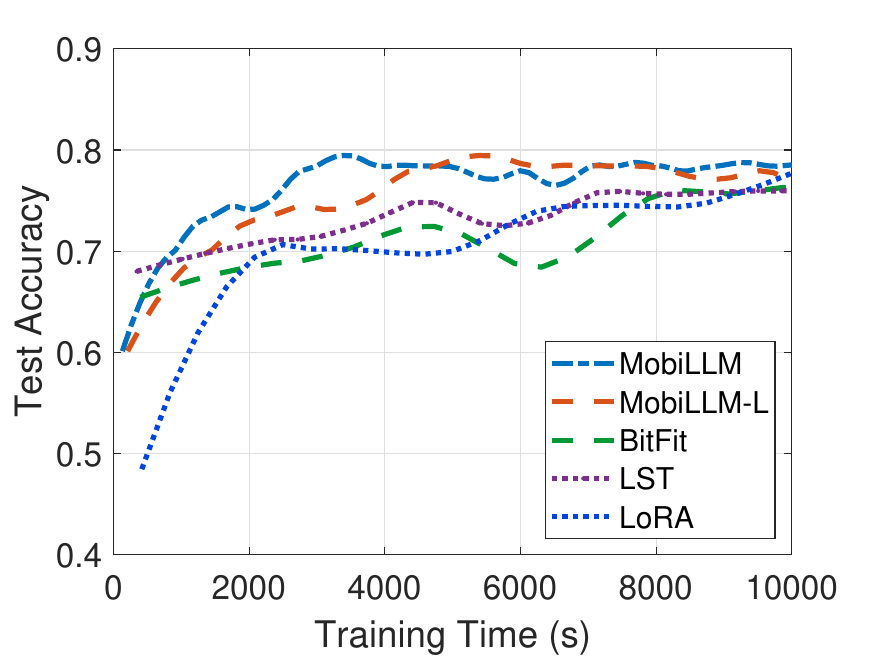}}
	\caption{Convergence performance on various models and tasks (On laptop). \label{Fig:ConvOPTlaptop}}
\end{figure*}

\textbf{MobiLLM accelerates on-device fine-tuning.} Figure~\ref{Fig:ConvOPTXavier} and Fig.~\ref{Fig:ConvOPTlaptop} show training curves that evaluate the accuracy-to-time performance of MobiLLM and other baselines. Specifically, the results in Fig.~\ref{Fig:ConvOPTXavier} are obtained by fine-tuning OPT-350M/1.5B models on a Xavier device equipped with a GPU offering 4.6 GB DRAM for model training. Figure~\ref{Fig:ConvOPTlaptop} shows results from experiments on a CPU-only laptop, which, unlike Xavier, lacks hardware acceleration for neural network computations but benefits from larger available memory. Here, we exclude the results of the Full-FT method as its memory requirements exceed the capabilities of both devices.
	
Overall, the proposed MobiLLM consistently outperforms its peer designs across various LLMs and tasks, achieving significant training speedups while maintaining comparable accuracy. Compared with the baseline, LoRA, MobiLLM expedites the fine-tuning to the target test accuracy by an average of $1.8\times$ on Xavier and $2.3\times$ on the laptop. Particularly, as shown in Fig.~\ref{Fig:ConvOPTXavier}, only MobiLLM can support training OPT-1.3B models on the Xavier device. As discussed in Sec.~\ref{sec:procedure}, unlike LoRA and BitFit, which require the calculation of gradients on frozen backbone parameters, MobiLLM only updates the side-network. This avoids the extensive computational costs of performing backpropagation on the backbone model, which ultimately expedites fine-tuning. Notably, MobiLLM exhibits superior performance to LST. LST employs transformer-based side networks. Despite the reduction of layer dimensions, transformer-based side-networks in LST still suffer from higher FLOPS than our adapter-based design due to compute-intensive mechanisms and feed-forward layers in the transformer architecture. Since the mobile device in MobiLLM focus solely on inference-related computations, cutting-edge inference-centric hardware acceleration techniques can be effortlessly integrated to further achieve speedups. In addition, the high-performance server accelerates forward and backward computations for the side-network, which also speeds up the training. 

\subsection{Sensitivity Study}
\label{subsec:Sensitivity Study}

\textbf{Sensitivity to activation quantization levels.} Table~\ref{table:CompareQuanti} illustrates how varying quantization levels of intermediate activation values influence the performance of MobiLLM. The results show that MobiLLM achieves similar accuracy performance across all quantization settings. When adopting the FP4 data type, MobiLLM incurs an average accuracy drop of only 1\% compared to non-quantized methods while reducing per-iteration transmission volume by approximately $4\times$. When adopting the NP4 data format, the accuracy remains even more stable. The results validate that MobiLLM can balance the trade-off between fine-tuning accuracy and transmission burden by appropriately selecting the activation quantization level. Furthermore, MobiLLM is compatible with diverse quantization techniques and benefits from their performance gains, which allows MobiLLM to tolerate quantization noise within an acceptable range to achieve the desired fine-tuning performance.
	
\textbf{Sensitivity to transmission rates.} By fixing the activation quantization level to be 4-bit, we also study the impacts of different transmission rates on the convergence time of MobiLLM, as reported in Table~\ref{table:SensRate}. To ensure the side-network effectively learns from the device's data, the mobile device occasionally reports its intermediate activations, whose size is proportional to the product of batch size and sequence length. At relatively higher uplink rates (e.g., 60 Mbps and 100 Mbps), MobiLLM converges faster than the device-only method despite introducing transmission delays. This is achieved through our overlapping training and activation quantization designs, which allow the mobile device to continuously conduct forward propagation concurrently with low-bit activation transmission. Moreover, the quantization of activations effectively reduces the transmission delays. The server’s efficient side-network computation further accelerates training compared to using a mobile device alone, mitigating the impact of transmission delays. Even at a slow transmission rate (e.g., 10 Mbps), MobiLLM maintains acceptable training time (only a 7\% increase compared to device-only), where the transmission delay emerges as a bottleneck. Notably, with a batch size larger than 24, MobiLLM can still support on-device fine-tuning, although the longer transmission delay slightly prolongs the training time. In contrast, device-only methods exceed the device's memory capacity and fail to perform fine-tuning under those settings. Future deployment or wide coverages of high-speed networks (e.g., 5G, 6G, or Wi-Fi 6) will strengthen MobiLLM’s power to enable or accelerate LLM fine-tuning on resource-constrained mobile devices.

\begin{table}[]
	\captionsetup{justification=centering}
	\caption{MobiLLM performance comparison with different activation quantizers. (Batch size=16, Sequence length=256)}
	\label{table:CompareQuanti}
	\small
	\setlength{\tabcolsep}{4pt}
	\vspace{-2mm}
	\begin{tabular}{c|c|c|c}
\hline
\multirow{2}{*}{Model} &
  \multirow{2}{*}{\begin{tabular}[c]{@{}c@{}}Quantizer\\ Configuration\end{tabular}} &
  \multirow{2}{*}{\begin{tabular}[c]{@{}c@{}}Data \\ Size (MB)\end{tabular}} &
  \multirow{2}{*}{Accuracy} \\
                          &                 &       &      \\ \hline
\multirow{4}{*}{OPT-350M} & No Act. Quant.  & 190   & 79.1 \\ \cline{2-4} 
                          & FP8 Act. Quant. & 99.6  &    78.6  \\ \cline{2-4} 
                          & FP4 Act. Quant. & 49.2  & 78.1 \\ \cline{2-4} 
                          & NP4 Act. Quant. & 49.2  & 78.8 \\ \hline
\multirow{4}{*}{OPT-1.3B} & No Act. Quant.  & 400   & 81.5 \\ \cline{2-4} 
                          & FP8 Act. Quant. & 200.3 &    81.2  \\ \cline{2-4} 
                          & FP4 Act. Quant. & 100.2 & 80.4 \\ \cline{2-4} 
                          & NP4 Act. Quant. & 100.2 & 81.3 \\ \hline
\end{tabular}
\end{table}

\begin{table}[]
	\captionsetup{justification=centering}
	\caption{MobiLLM sensitivity with different transmission rate (Model: OPT-350M).}
	\label{table:SensRate}
	\small
	\setlength{\tabcolsep}{4.5pt}
	\vspace{-2mm}
	\begin{tabular}{c|c|c|c|c}
\hline
\begin{tabular}[c]{@{}c@{}}Time Per \\ Iteration (s)\end{tabular} & 10Mbps & 60Mbps & 100Mbps & \begin{tabular}[c]{@{}c@{}}Device\\ only\end{tabular} \\ \hline
Batch size=12 & 7.3  & 5.5  & 5.4  & 6.8 \\ \hline
Batch size=16 & 9.8  & 7.48 & 7.5  & 8.4 \\ \hline
Batch size=24 & 14.7 & 10.1 & 10   & ---   \\ \hline
Batch size=30 & 18.4 & 14   & 14.1 & ---   \\ \hline
\end{tabular}
\end{table}

\section{Conclusion and Future Work}

In this paper, we have introduced MobiLLM, a novel framework for fine-tuning transformer-based LLMs on the mobile device. MobiLLM separates parallel adapters from the backbone, and offloads dominant computational and memory burdens in the fine-tuning process to the high-performance server. It allows the mobile device to retain a frozen backbone model and perform only forward propagation parallel to the server-side execution, while keeping their data locally on the device. Comparative analyses revealed that MobiLLM outperforms existing fine-tuning methods, including a $4\times$ reduction in memory footprint and a $2.3\times$ speedup in training convergence. MobiLLM thus unlocks the potential for fine-tuning billion-sized LLMs, such as OPT-1.3B, on resource-constrained mobile devices, even a CPU-only one.

As a future extension, MobiLLM could allow the parameters of the pre-trained backbone model to be quantized into low-precision values to further reduce the memory footprint of weights. This helps facilitate easier storage and deployment of large-sized LLMs on resource-constrained mobile devices. Our additional experiments reveal that with the backbone model quantized to INT4, MobiLLM limits on-device memory usage to 1.176 GB when fine-tuning OPT-350M (batch size=16, sequence length=256). For OPT-1.3B, the memory usage is remarkably reduced to 2.751 GB — over a 5x reduction compared to LoRA fine-tuning — while maintaining a negligible accuracy drop of just 2\%. Besides, low-precision data are typically faster to execute on modern accelerators by utilizing the integer kernels, which are supported by a wide range of hardware (e.g., NVIDIA GPUs, Intel CPUs, Qualcomm DSPs, etc.). As a next step, we will explore the integration of MobiLLM with advanced post-training quantization techniques to further achieve hardware acceleration while preserving model accuracy within tight memory constraints. We will also design a task-oriented activation filtering and transmission mechanism to selectively transmit important activations to side networks for further reducing transmission overhead.

\bibliographystyle{ACM-Reference-Format}
\bibliography{Liang}
\clearpage
\end{document}